\documentclass[sigconf,natbib,screen=true]{acmart}
\usepackage{amsmath}
\usepackage{booktabs} 
\usepackage{algorithm, algpseudocode}
\usepackage{graphics}
\usepackage{epsfig}
\usepackage{graphicx}
\usepackage{xcolor}
\usepackage[skip=0pt]{caption}
\usepackage{subfigure}
\usepackage{balance}
\usepackage{multirow}
\usepackage{mathrsfs}
\usepackage{acronym}
\usepackage{placeins}
\usepackage{xcolor}
\usepackage[inline]{enumitem}
\usepackage{fancyhdr}
\usepackage{tabularx}
\usepackage{bm}

\AtBeginDocument{%
  \providecommand\BibTeX{{%
    \normalfont B\kern-0.5em{\scshape i\kern-0.25em b}\kern-0.8em\TeX}}}

\acrodef{CRSs}{conversational recommender systems}
\acrodef{CRS}{conversational recommender system}
\acrodef{mdn}{my model name}
\acrodef{KET}{graphs embedding with external text}
\acrodef{KGs}{knowledge graphs}
\acrodef{KG}{knowledge graph}
\acrodef{CR}{conversational recommendation}

\newcommand{\OurMethod}{VRICR}
\newcommand{\header}[1]{\vspace*{1mm}\noindent\textbf{#1.}}

\copyrightyear{2023}
\acmYear{2023}
\setcopyright{acmlicensed}\acmConference[WSDM '23]{Proceedings of the
Sixteenth ACM International Conference on Web Search and Data
Mining}{February 27-March 3, 2023}{Singapore, Singapore}
\acmBooktitle{Proceedings of the Sixteenth ACM International Conference on
Web Search and Data Mining (WSDM '23), February 27-March 3, 2023,
Singapore, Singapore}
\acmPrice{15.00}
\acmDOI{10.1145/3539597.3570426}
\acmISBN{978-1-4503-9407-9/23/02}
\begin{document}
\fancyfoot{}


\author{Xiaoyu Zhang}
\affiliation{
  \institution{Shandong University}
  \city{}
  \country{}
}
\email{xiaoyu.zhang@mail.sdu.edu.cn}

\author{Xin Xin}
\affiliation{
  \institution{Shandong University}
  \city{}
  \country{}
}
\email{xinxin@sdu.edu.cn}

\author{Dongdong Li}
\affiliation{
  \institution{Shandong University}
  \city{}
  \country{}
}
\email{lddsdu@gmail.com}

\author{Wenxuan Liu}
\affiliation{
  \institution{Shandong University}
  \city{}
  \country{}
}
\email{wenxuan.liu@mail.sdu.edu.cn}

\author{Pengjie Ren}
\affiliation{
  \institution{Shandong University}
  \city{}
  \country{}
}
\email{jay.ren@outlook.com}

\author{Zhumin Chen}
\affiliation{
  \institution{Shandong University}
  \city{}
  \country{}
}
\email{chenzhumin@sdu.edu.cn}

\author{Jun Ma}
\affiliation{
  \institution{Shandong University}
  \city{}
  \country{}
}
\email{majun@sdu.edu.cn}

\author{Zhaochun Ren}
\affiliation{
  \institution{Shandong University}
  \city{}
  \country{}
}
\authornote{corresponding author.}
\email{zhaochun.ren@sdu.edu.cn}

\renewcommand{\shortauthors}{Xiaoyu Zhang et al.}

\title{Variational Reasoning over Incomplete Knowledge Graphs \\for Conversational Recommendation}

\begin{abstract}
\Acf{CRSs} often utilize external \acf{KGs} to introduce rich semantic information and recommend relevant items through natural language dialogues. However, original \ac{KGs} employed in existing \ac{CRSs} are often incomplete and sparse, which limits the reasoning capability in recommendation. Moreover, only few of existing studies exploit the dialogue context to dynamically refine knowledge from \ac{KGs} for better recommendation.
To address the above issues, we propose the \textbf{V}ariational \textbf{R}easoning over \textbf{I}ncomplete
KGs \textbf{C}onversational \textbf{R}ecommender (VRICR).
Our key idea is to 
incorporate the large dialogue corpus naturally accompanied with \ac{CRSs} to enhance the incomplete \ac{KGs}; and
perform dynamic knowledge reasoning conditioned on the dialogue context.
Specifically, we denote the dialogue-specific subgraphs of \ac{KGs} as latent variables with categorical priors for adaptive knowledge graphs refactor. We propose a variational Bayesian method to approximate posterior distributions over dialogue-specific subgraphs, which not only leverages the dialogue corpus for restructuring missing entity relations but also dynamically selects knowledge based on the dialogue context.
Finally, we infuse the 
{dialogue-specific subgraphs} to decode the recommendation and responses. We conduct experiments on two benchmark \ac{CRSs} datasets. Experimental results confirm the effectiveness of our proposed method.

\end{abstract}



\maketitle

\acresetall


\section{Introduction}

\begin{figure}
    \centering
    \includegraphics[width=\linewidth]{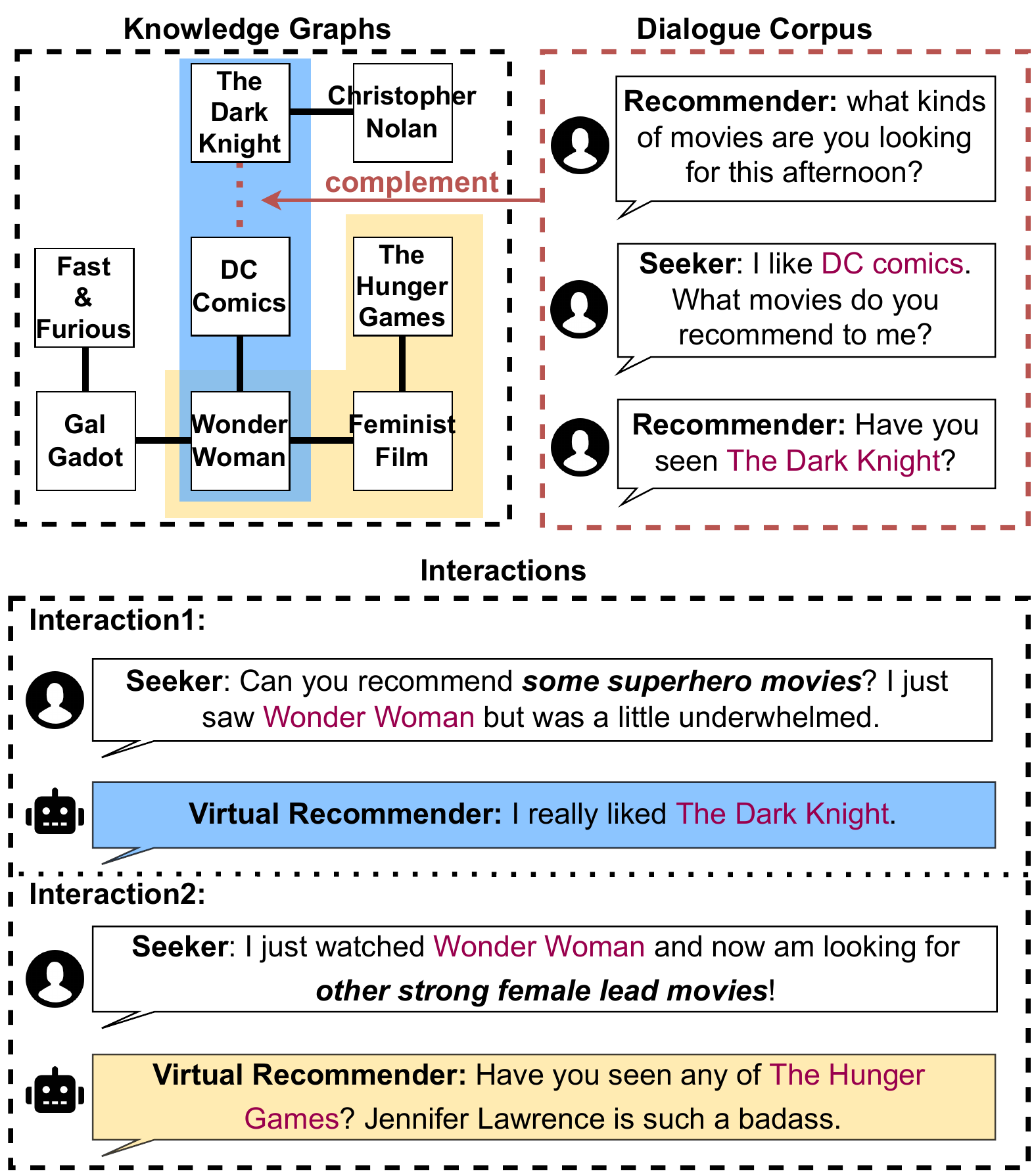}
    \caption{ For \emph{Interaction1}, the relation between ``The Dark Knight'' and ``DC Comics'' is missing in the original KGs. By incorporating the dialogue corpus, the missing relation (red dotted line) can be completed and reused for recommendation in \emph{Interaction1}. For  \emph{Interaction1} and \emph{Interaction2}, correct knowledge (blue shaded area and yellow shaded area, separately) should be selected conditioned on the dialogue context (bold italics text).}
    \label{fig:incomplete graph examples}
\end{figure}

Unlike classic recommender systems, \acf{CRSs} recommend items to users more interactively through multi-turn dialogues.
Recent studies on CRSs focus on implicitly inferring user interests through the understanding of natural interaction utterances~\citep{DBLP:conf/nips/LiKSMCP18,DBLP:conf/emnlp/ChenLZDCYT19,DBLP:conf/kdd/ZhouZBZWY20,DBLP:conf/emnlp/ZhouWHH21,DBLP:conf/wsdm/ZhouZZWJ022}.
To help recommender systems comprehend semantics of complicated dialogues, external knowledge, e.g., \ac{KGs}, has been involved for \ac{CRSs}, which successfully improves the performance of conversational recommendation~\citep{DBLP:conf/emnlp/ChenLZDCYT19,DBLP:conf/kdd/ZhouZBZWY20,DBLP:conf/emnlp/ZhouWHH21,DBLP:conf/wsdm/ZhouZZWJ022}. 

However, incomplete, sparse, and ambiguous information in \ac{KGs} still limit the further exploitation of \ac{KGs} for \ac{CRSs} in the following two folds~\cite{DBLP:conf/aaai/MalaviyaBBC20,DBLP:conf/www/0003W0HC19,DBLP:conf/coling/SarkarGAM20}:
(1) Information missed in the incomplete and sparse \ac{KGs} forms multiple independent subgraphs, which reduces the reasoning ability of \ac{CRSs}. Traditional \ac{KGs} completion methods, both path-based and embedding-based, encounter difficulties in these \ac{KGs} with independent subgraphs because of the lack of essential paths (for path-based methods) or structural information (for embedding-based methods) during ineffective learning~\citep{DBLP:conf/naacl/ChenXYW18,DBLP:conf/nips/BordesUGWY13,DBLP:journals/corr/YangYHGD14a}.
As shown in Fig.~\ref{fig:incomplete graph examples}, we see solid lines indicate relations existing in the original \ac{KGs}, while the bottom half of the \ac{KGs} forms an independent subgraph without the entity ``The Dark Knight''. 
This unconnected subgraph devastates the \emph{Interaction1} conversation as ``The Dark Knight'' cannot be inferred given the original \ac{KGs}.
(2) Existing studies which simply utilize entire \ac{KGs} for various interactions cannot adaptively select proper knowledge for personalized recommendation~\citep{DBLP:conf/nips/LiKSMCP18,DBLP:conf/emnlp/ChenLZDCYT19,DBLP:conf/kdd/ZhouZBZWY20,DBLP:conf/emnlp/ZhouWHH21,DBLP:conf/wsdm/ZhouZZWJ022}.
As shown in the bottom part of Fig.~\ref{fig:incomplete graph examples}, the bold italics text in two conversations (i.e., \emph{Interaction1} and \emph{Interaction2}) represents different dialogue contexts although both of them mention ``Wonder Woman''. 
Thus for each conversation, instead of selecting the same knowledge from the entire graph, an appropriate response should be generated based on a specific subgraph in \ac{KGs} including the entity ``Wonder Woman'' and other entities/relations relevant to the corresponding context.
How to perform dynamic knowledge reasoning conditioned on the dialogue context to select relevant knowledge is still an open question in unsupervised paradigms.

To overcome the above limitations, we situate our study in \ac{CRSs} with dynamic knowledge reasoning over incomplete \ac{KGs}. 
Intuitively, we believe incorporating the evidence from both dialogue corpus and \ac{KGs} helps
\begin{enumerate*}[label=(\roman*)] 
    \item complement missing but useful knowledge links in \ac{KGs}; and 
    \item perform adaptive knowledge reasoning over \ac{KGs}  for personalized recommendation.
\end{enumerate*}
As shown in Fig.~\ref{fig:incomplete graph examples}, we aim to explicitly infer the missing link between ``The Dark Knight'' and ``DC Comics'' (i.e., the red dashed line), through incorporating the dialogue corpus. Then the inferred relation will be reused to enhance the recommendation in \emph{Interaction1}, as the blue shaded area shows. 
Besides, as shown in \emph{Interaction2}, we aim to dynamically choose relevant knowledge for specific interactions conditioned on the dialogue context, as the yellow shaded area shows.

To this end, we propose the \emph{Variational Reasoning over Incomplete \ac{KGs} Conversational Recommender}  (\OurMethod{}).
\OurMethod{} denotes the dialogue-specific subgraphs of KGs as discrete latent variables which will be inferred in a variational Bayesian manner~\citep{DBLP:journals/corr/KingmaW13}, recommendation as the observed variable, dialouge context and the original \ac{KGs} as the conditions. 
We frame \OurMethod{} as four main modules: the knowledge graph refactor network, the context encoder, the recommender, and the response generator. The knowledge graph refactor network includes the prior and approximate posterior networks, whereas the recommender is viewed as the likelihood network. 
From this perspective, we employ a stochastic gradient variational Bayes (SGVB) estimator to optimize the derived evidence lower bound. 
As a result, the knowledge graph refactor network synchronously infers the missing relations of \ac{KGs} and select knowledge given the output of the context encoder. Then we fuse the output of the knowledge graph refactor network, i.e., the dialogue-specific subgarphs, to the recommender and the response generator to accomplish the \acf{CR} task.

The contributions of this paper are as follows:
\begin{enumerate*}[label=(\roman*)]
    \item 
    Our work is the first attempt to explicitly address the problem of dynamic reasoning over incomplete \ac{KGs} for \ac{CRSs}.
    \item We incorporate the large dialogue corpus naturally accompanied with \ac{CRSs} to enhance the incomplete \ac{KGs}.
    \item We propose a variational reasoning approach that dynamically refines and utilizes relevant knowledge in an unsupervised fashion. 
    \item Experimental results show that our proposed \OurMethod{} achieves significant improvement over baselines in the \ac{CR} task. 
    
\end{enumerate*}
\vspace{-2mm}
\section{Related work}
\label{sec:related work}
\subsection{ Graphs embedding with text}
One line of research related to our research problem utilizes external textual information to enhance the  graph embedding. 
Several previous work~\citep{DBLP:conf/ijcai/WangL16,DBLP:conf/emnlp/KartsaklisPC18} enriches graphs based on the occurrence of entities in external text. \citet{DBLP:conf/ijcai/WangL16} develop a neural network based on entity-word co-occurrences. \citet{DBLP:conf/emnlp/KartsaklisPC18} add weighted edges between entities and textual features, and then learn graphs representations using random walks. 
Other studies propose learning-based methods for graphs enrichment~\citep{DBLP:conf/emnlp/SunDZMSC18,DBLP:conf/aaai/MalaviyaBBC20,DBLP:conf/naacl/RezayiZKRLL21}. 
\citet{DBLP:conf/emnlp/SunDZMSC18} apply GCN~\citep{DBLP:conf/iclr/KipfW17} to extract subgraphs, which integrate the semantics of graphs and text sentences for question-answering tasks. 
\citet{DBLP:conf/aaai/MalaviyaBBC20} propose to use transfer learning from pretrained language models to graph representations. 
\citet{DBLP:conf/naacl/RezayiZKRLL21} create augmented graphs from an external source and improve the quality of node representations by jointly modeling original and augmented graphs.

\ac{CRSs} are naturally accompanied by a large dialogue corpus. We migrate the task of graphs embedding with external text to \ac{CRSs}. Our model is capable of enhancing the \ac{KGs} by fully promoting the semantic fusion between the dialogue corpus and original \ac{KGs}, which leads to overcoming the incompleteness and sparsity in \ac{KGs} for better recommendation and response generation.

\subsection{Conversational recommender systems}
\begin{figure*}
    \centering
    \includegraphics[width=1.0\linewidth]{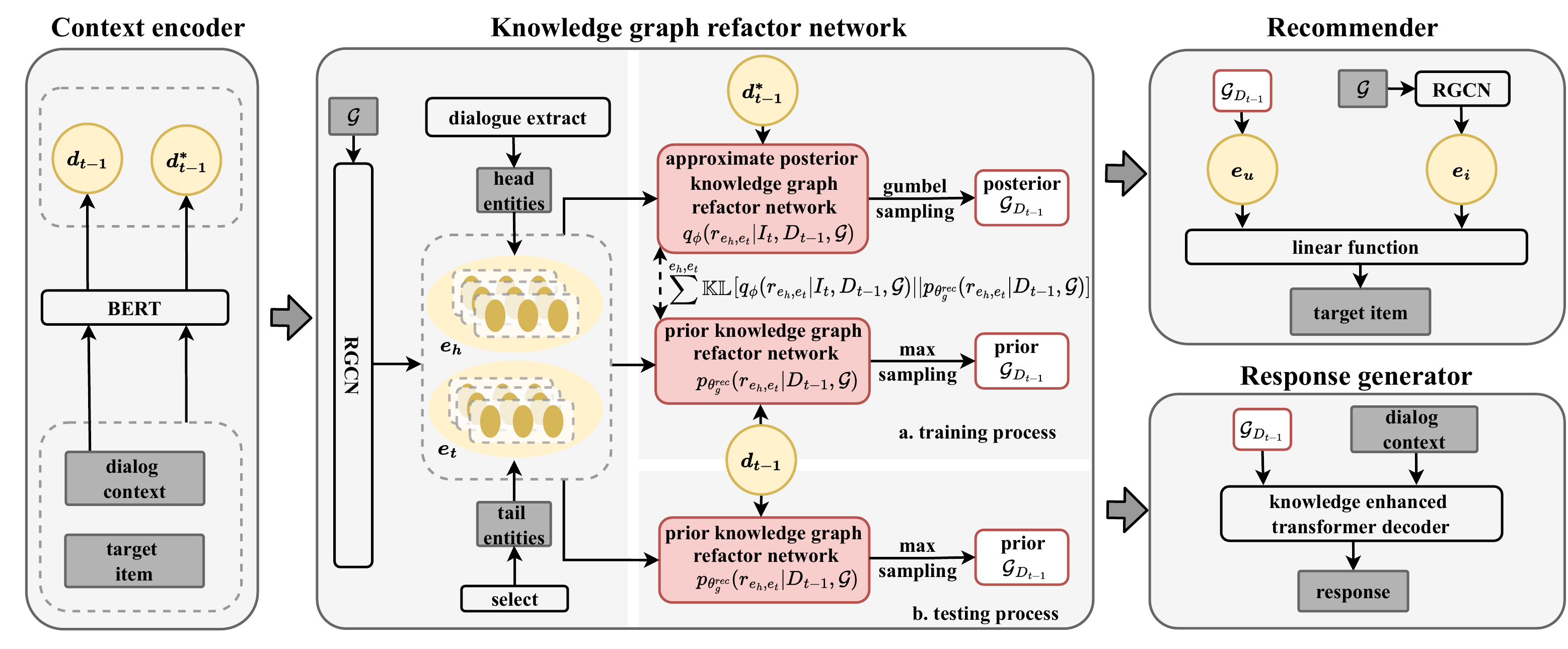}
    \caption{An overview of  \OurMethod{}. \OurMethod{} has four components: the context encoder, the knowledge graph refactor network, the recommender and the response generator.}
    \label{fig:framework}
\end{figure*}
According to~\citet{DBLP:journals/aiopen/GaoLHRC21}, \acf{CRSs} can be divided into attribute-based and topic-guided models.

\textbf{Attribute-based \ac{CRSs}} focus on the recommendation strategy in \acp{CRS}, including ``whe\-ther to ask or recommend'', ``which attributes to ask'' or ``which items to recommend.'' 
Lots of attribute-based \ac{CRSs} use deep reinforcement learning framework to select appropriate attributes or items through multi-turns interactions~\citep{DBLP:conf/sigir/SunZ18,DBLP:conf/wsdm/Lei0MWHKC20}. 
Recently, \citet{DBLP:conf/sigir/DengL0DL21} and \citet{DBLP:conf/kdd/LeiZ0MWCC20} use \ac{KGs} derived from external data sources to improve the recommendation performance.

\textbf{Topic-guided \ac{CRSs}} focus on interacting with users through natural language conversations, emphasizing fluent response generation and precise recommendation.
Unlike attribute-based \ac{CRSs}, Topic-guided \ac{CRSs} provides considerable flexibility in influencing how the conversation continues.
\citet{DBLP:conf/nips/LiKSMCP18} utilize an auto-encoder~\citep{DBLP:conf/www/SedhainMSX15} for recommendation and a hierarchical RNN for response generation.
\citet{DBLP:conf/emnlp/ChenLZDCYT19} propose an end-to-end framework for conversational recommendation using external knowledge. 
\citet{DBLP:conf/acl/LiuWNWCL20} propose a multi-goal driven conversation generation framework to lead conversations to recommendation proactively. 
\citet{DBLP:conf/emnlp/ZhangYC0Y21} consider more accurate knowledge retrieval driven by subgoals prediction. \citet{DBLP:conf/kdd/ZhouZBZWY20} incorporate word-oriented and entity-oriented \ac{KGs}. 
Based on that, \citet{DBLP:conf/emnlp/ZhouWHH21} apply reinforcement learning to conduct multi-hop reasoning on KGs. \citet{DBLP:conf/emnlp/MaTH21} perform tree-structured reasoning on a \ac{KGs}.
\citet{DBLP:conf/wsdm/ZhouZZWJ022} propose a coarse-to-fine contrastive learning method to improve data semantic fusion.
\citet{DBLP:conf/kdd/WangZWZ22} unify the recommendation and conversation subtasks into the prompt learning paradigm and utilizes a pretrained language model.
Recently, another kind of Topic-guided \ac{CRSs} has been proposed, which leverages users' historical interactions for preference capturing in the conversational recommendation.
\citet{DBLP:conf/sigir/RenTLR0XLRC22} propose an end-to-end variational reasoning approach by jointly modeling both long-term preferences and short-term preferences as latent variables.
\citet{DBLP:conf/sigir/ZouKRRSL22} construct a sequence of items mentioned in the dialogue context, and discover user preferences by modeling the dependencies. 
\citet{DBLP:conf/sigir/LiXZ0Z022} highlight the importance of users’ historical dialogue sessions and look-alike users for multi-aspect preference learning. 

Recent studies which fuse semantic information of the \ac{KGs} in \ac{CRSs}~\citep{DBLP:conf/emnlp/ChenLZDCYT19,DBLP:conf/kdd/ZhouZBZWY20,DBLP:conf/emnlp/ZhouWHH21,DBLP:conf/wsdm/ZhouZZWJ022,DBLP:conf/emnlp/MaTH21,DBLP:conf/kdd/WangZWZ22} still faces two challenging problems: (\romannumeral1) incompleteness and sparsity in the \ac{KGs}; and (\romannumeral2) extraneous information with noisy signals in the \ac{KGs}. They simply incorporate the entire incomplete \ac{KGs} for different interactions. In this paper, to deal with incompleteness and sparsity, we focus on explicitly inferring relations that do not originally exist in the KGs. And to reduce extraneous information, we construct dialogue-specific subgraphs for knowledge refinement in \ac{CRSs}.
\section{Method}
\label{sec:method}
In this section, we first formulate our research problem in Sec.\ref{subsec:problem formulation}. Then we introduce our variation Bayesian paradigm in Sec.~\ref{subsec:vbm}. Based on the Bayesian model, the proposed  \OurMethod{} has four main components: the context encoder (Sec.~\ref{subsec:context encoder}), the knowledge graph refactor network (Sec.~\ref{subsec:KG Refactor network}), the recommender (Sec.~\ref{subsec:Recommender System}) and the response generator (Sec.~\ref{subsec:Dialouge System}). Fig.~\ref{fig:framework} shows an overview of \OurMethod{}.

\subsection{Problem formulation}
\label{subsec:problem formulation}

\textbf{Notations}. Given ${t-1}$ dialogue turns, the dialogue context ${D_{t-1}}$ consists of a sequence of utterances from the user or the conversational recommender, i.e., ${D}_{t-1} = \{{U_i}\}^{t-1}_{i=1}$. 
Each utterance ${U}_{i}=\{{w}_{j}\}^{|{U}_{i}|}_{j=1}$ consists a sequence of words. 
At the $t$-th turn, the recommender chooses a set of target items $I_t =  \{{I_i}\}^{|{I_t}|}_{i=1}$ for the user. 
We denote $\mathcal{G}=(\mathcal{E},\mathcal{R})$ as an external \ac{KG} consisting of triples $<e_{h}, r, e_{t}>$, where $e_h \in \mathcal{E}$ and $e_{t}\in \mathcal{E}$ are the head and tail entities, $r \in \mathcal{R}$ reflects the relation between $e_h$ and $e_t$. 
We assume that all items are included in vertices of $\mathcal{G}$, i.e., $I_i \in \mathcal{E}$.

Given the dialogue context ${D_{t-1}}$, there is a dialogue-specific subgraph $\mathcal{G}_{D_{t-1}}$.  $\mathcal{G}_{D_{t-1}}$ adaptively retains correlated entities and relations in $\mathcal{G}$ conditioned on ${D_{t-1}}$.
Besides, due to the fact that  $\mathcal{G}$ is incomplete\cite{DBLP:conf/www/0003W0HC19}, $\mathcal{G}_{D_{t-1}}$ additionally introduce missing but useful relations through the knowledge graph refactor network to better accomplish the \ac{CR} task.

We consider our model with parameters $\theta^{rec}$ and $\theta^{dia}$ for recommendation and response generation subtasks, correspondingly. More precisely, $\theta^{rec}$ contains the parameters of the knowledge graph refactor network as $\theta^{rec}_g$, and the recommender as $\theta^{rec}_r$.

\noindent
 \textbf{Task Definition}. The task of \ac{CR} is formulated as follows: in the training stage, given the incomplete graph $\mathcal{G}$ and dialogue context $D_{t-1}$, we target to learn $\theta^{rec}$ and  $\theta^{dia}$ through maximizing the probability distribution over observed $I_t$ and ${U}_{t}$ in the training data, so we have:
\begin{equation}
\label{tgt_rec}
\begin{aligned}
      &\theta^{rec}=\text{argmax }P_{\theta^{rec}}(I_t|D_{t-1},\mathcal{G}),\\ &\theta^{dia}=\text{argmax }P_{\theta^{dia}}(U_{t}|D_{t-1},\mathcal{G}).
\end{aligned}
\end{equation}
In the inference stage, we aim to use the model with parameters  $\theta^{rec}$ and  $\theta^{dia}$ to (1) accurately recommend items; and (2) generate proper responses to users.

\subsection{Variational Bayesian paradigm}
\label{subsec:vbm}

\begin{figure}
    \centering
    \includegraphics[width=0.7\linewidth]{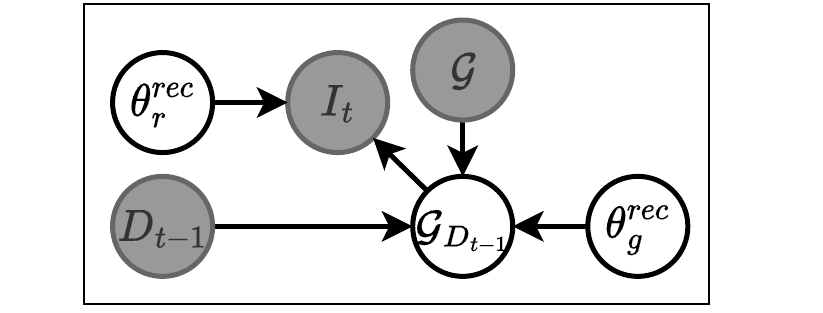}
    \caption{The graphical representation of  \OurMethod{}. Shaded nodes represent observed variables.}
    \Description{The framework of \OurMethod{}.}
    \label{fig:GPM}
\end{figure}

To perform dynamic reasoning conditioned on the dialogue context $D_{t-1}$ over the incomplete KG $\mathcal{G}$ for personalized recommendation, we regard the $\mathcal{G}_{D_{t-1}}$ as latent variables. 
The probability $P_{\theta^{rec}}(I_t|D_{t-1},\mathcal{G})$ in Eq.~\ref{tgt_rec} is then reformulated as follows:
\begin{equation}
\label{eq:Bayesian generative model}
\begin{aligned}
    &\sum_{\mathcal{G}_{D_{t-1}}}P_{\theta_{r}^{rec}}(I_{t}|\mathcal{G}_{D_{t-1}})\cdot P_{\theta_{g}^{rec}}(\mathcal{G}_{D_{t-1}}|D_{t-1},\mathcal{G}),
\end{aligned}
\end{equation}
where $P_{\theta_{r}^{rec}}(I_{t}|\mathcal{G}_{D_{t-1}})$ is derived using the recommender, and the probability distribution $P_{\theta_{g}^{rec}}(\mathcal{G}_{D_{t-1}}|D_{t-1},\mathcal{G})$ is estimated through the knowledge graph refactor network.
The graphical representation of  \OurMethod{} is shown in Fig.~\ref{fig:GPM}, where shaded and unshaded nodes indicate observed and latent variables, respectively. 

We introduce $I_t$ as additional supervision signals to estimate the posterior distribution $P(\mathcal{G}_{D_{t-1}}| I_t,{D_{t-1}},\mathcal{G})$. Since the exact posterior distribution is intractable, we utilize an approximate posterior knowledge graph refactor network parametered by $\phi$ to approximate the posterior distribution as 
$Q_{\phi}(\mathcal{G}_{D_{t-1}}| I_t,{D_{t-1}},\mathcal{G})$.
In the training stage, the prior knowledge graph refactor network is trained through a variational Bayesian manner, i.e., to minimize  the difference between $P_{\theta_{g}^{rec}}(\mathcal{G}_{D_{t-1}}|D_{t-1},\mathcal{G})$ and $Q_{\phi}(\mathcal{G}_{D_{t-1}}| I_t,{D_{t-1}},\mathcal{G})$, meanwhile to maximize the likelihood of observed recommendation. 

More precisely, we formulate the distribution  $P_{\theta_{g}^{rec}}(\mathcal{G}_{D_{t-1}}|D_{t-1},\mathcal{G})$ and $Q_{\phi}(\mathcal{G}_{D_{t-1}}| I_t,{D_{t-1}},\mathcal{G})$ as the multiplication among the probability of relations between entity pairs:
\begin{equation}
\label{eq:prior}
    \begin{aligned}
        P_{\theta_{g}^{rec}}(\mathcal{G}_{D_{t-1}}|D_{t-1},\mathcal{G}) &= \prod^{e_h ,e_t }p_{\theta_{g}^{rec}}(r_{{e_h},{e_t}}|D_{t-1},\mathcal{G}),\\
        Q_{\phi}(\mathcal{G}_{D_{t-1}}|I_t,{D_{t-1}},\mathcal{G}) &= \prod^{e_h ,e_t}q_{\phi}(r_{{e_h},{e_t}}|I_t,{D_{t-1}},\mathcal{G}),
    \end{aligned}
\end{equation}
where $p_{\theta_{g}^{rec}}(r_{{e_h},{e_t}}|D_{t-1},\mathcal{G})$ and $q_{\phi}(r_{{e_h},{e_t}}|I_t,{D_{t-1}},\mathcal{G})$ are the prior and approximate posterior probability distributions of relations, respectively.

\header{Evidence Lower Bound (ELBO)}
We derive the ELBO for optimizing the prior and approximate posterior knowledge graph refactor networks simultaneously as:
\begin{small}
\begin{equation}
\label{eq:ELBO}
\begin{aligned}
    && &P_{\theta^{rec}}(I_t|D_{t-1},\mathcal{G}),\\
    &&\ &\ge \mathbb{E}_{\prod^{e_h ,e_t} q_{\phi}(r_{e_h,e_t}|I_t,{D_{t-1}},\mathcal{G} )}
    \left[
    P_{\theta_{r}^{rec}}
        \left( {I}_{t}|\mathcal{G}_{D_{t-1}} \right) 
    \right]\\
    &&\  &\ \  \ \ - \sum^{e_h,e_t} \mathbb{KL}
    \left[  q_{\phi} \left( r_{{e_h},{e_t}} |I_t,{D_{t-1}},\mathcal{G}  \right) ||p_{\theta_{g}^{rec}}\left(r_{{e_h},{e_t}}|{D_{t-1}},\mathcal{G}  \right) \right], \\
\end{aligned}
\end{equation}
\end{small}
\noindent where $ \mathbb{E}(\cdot) $ is the expectation, and $ \mathbb{KL}(\cdot || \cdot) $ denotes the Kullback-Leibler divergence. To estimate Eq.~\ref{eq:ELBO}, we sample the relation between the entity pair $(e_h,e_t)$ from $q_{\phi}(r_{{e_h},{e_t}}|I_t,{D_{t-1}},\mathcal{G})$ one by one and replenish the inferred relations to $\mathcal{G}_{D_{t-1}}$. 
Finally, $P_{\theta_{r}^{rec}} (I_{t}|{D_{t-1}},\mathcal{G})$ selects $I_{t}$ depending on $\mathcal{G}_{D_{t-1}}$. The above sampling procedure is shown in Fig.~\ref{fig:framework} (a. training process). 

\subsection{Context encoder}
\label{subsec:context encoder}
At the $t$-th turn, we use the BERT~\citep{DBLP:conf/naacl/DevlinCLT19} to get the representation $\boldsymbol{d}_{t-1}$ of dialogue context $D_{t-1}$. 
$\boldsymbol{d}_{t-1}$ is the condition input vector of the prior knowledge graph refactor network. 
Next, we concatenate the dialogue context $D_{t-1}$ and the word sequence of target item $I_t$, segmented by a special token, as the input of the same BERT encoder to learn a vector, i.e., $\boldsymbol{d}_{t-1}^{*}$,  which is the input vector of the approximate posterior knowledge graph refactor network.
It is worth noting that we use the word sequence of $I_t$ rather than the entity representation from the KG to encode the item information. Since we find the utilization of entity representation in the approximate posterior knowledge graph refactor network may lead to the graph semantic bias that the prior network potentially fails to follow the approximate posterior network.

\subsection{Knowledge graph refactor network}
\label{subsec:KG Refactor network}
\subsubsection{Graph encoding}
\label{subsubsec:Graph Encoding}
We adopt RGCN ~\citep{DBLP:conf/iclr/KipfW17} as the backbone for capturing the crucial relation semantics between entities. 
To facilitate the learning process, we use a pre-recommendation task to pretrain the representation $\boldsymbol{e} \in \mathbb{R}^d$ of the entity  $e$ 
in the RGCN.
More precisely, we stack the representations of entities appearing in the dialogue context $D_{t-1}$ into a matrix $\boldsymbol{\mathcal{N}}$. And we utilize a self-attentive mechanism to generate the user representation $\boldsymbol{u}$ in the pre-recommendation task as:
\begin{equation}
\begin{aligned}
\boldsymbol{u} &=\boldsymbol{\mathcal{N}} \cdot \boldsymbol{a}, \\
\boldsymbol{a} &=\operatorname{softmax}\left(\boldsymbol{b}^{\top} \cdot \tanh \left(\mathbf{W}_{\boldsymbol{a}} \boldsymbol{\mathcal{N}}\right)\right),
\end{aligned}
\end{equation}
where $\mathbf{W}_{\boldsymbol{a}}$ and $\boldsymbol{b}$ are parameters to learn.
Then we compute the probability of recommending an item $I_i$ to the user as:
\begin{equation}
\begin{aligned}
P_{\theta_{r}^{pre}}(I_i| {D_{t-1},\mathcal{G}}) = & \operatorname{softmax}\left(\boldsymbol{u}^{\top} \boldsymbol{e}_{i}\right),\\
\end{aligned}
\end{equation}
where $\boldsymbol{e}_{i}$ is the representation for item $I_i$ outputted by RGCN. $\theta_r^{pre}$ denotes the pretrained recommendation parameter. Finally, we apply a cross-entropy loss as the optimization objective for the pre-recommendation task:
\begin{equation}
\begin{aligned}
\mathcal{L}^{pre} =  &-\log P_{\theta_{r}^{pre}}(I_t | {D_{t-1}},\mathcal{G}).
\end{aligned}
\end{equation}
In this way, the knowledge graph refactor network is more adapted to the downstream recommendation subtask.

\subsubsection{Entities selection}
\label{subsubsec:Subgraph Entities Selection}

To improve efficiency and identify more important entities, we assume only entities appearing in the dialogue context can be head entities in $\mathcal{G}_{D_{t-1}}$.
We include two sources of tail entities. 
 
(1) We utilize the mutual information to measure the correlations of entities in $\mathcal{E}$:
\begin{equation}
\begin{aligned}
    P_c(e) = \sum_{e_h}P(e|e_h)P(e_h)
    \log \left[ \frac{P(e|e_h)}{P(e)} \right],
\end{aligned}
\end{equation}
where $P(e)$ is occurrence probability of an entity $e$ and $P(e|e_h)$ is co-occurrence probability of the entity $e$ given $e_h$, which are both counted from the training dataset. We choose the top-$k$ entities as a part of selected tail entities, where $k$ is a hyperparameter.

(2) We keep the tail entities that are related to the head entities in the original incomplete knowledge graph $\mathcal{G}$.

\subsubsection{Relation refactor}
\label{subsubsec:Relation Refactor}

Given the entity set of the subgraph and the entity representations, 
we formulate relation refactor as a classification task. The prior and approximate posterior knowledge graph refactor networks are both based on multilayer perceptron (MLP) ~\citep{gardner1998artificial}.
In order to fully integrate the semantics from head and tail entities as well as the dialogue context, given the entity pair and the context representation $\boldsymbol{e_h} , \boldsymbol{e_t} , \boldsymbol{d_{t-1}}$,  we denote the fused input of MLP in the prior knowledge graph refactor network in Eq.~\ref{Eq:info_fusion}:
\begin{equation}
\label{Eq:info_fusion}
    \begin{aligned}
        \boldsymbol{f^{o 1}}&=\left[\boldsymbol{e_h};\boldsymbol{e_t};\boldsymbol{d_{t-1}}\right], \\
        \boldsymbol{f^{o 2}}&=\left[\boldsymbol{e_h} \odot \boldsymbol{e_t};\boldsymbol{e_t} \odot \boldsymbol{d_{t-1}};\boldsymbol{e_h} \odot \boldsymbol{d_{t-1}}\right],\\
        \boldsymbol{f^{o 3}}&=\boldsymbol{e_h} \odot  \boldsymbol{e_t} \odot \boldsymbol{d_{t-1}}, \\
        \boldsymbol{m}&=\left[\boldsymbol{f^{o 1}};\boldsymbol{f^{o 2}}; \boldsymbol{f^{o 3}}\right] \mathbf{W}_{o},
    \end{aligned}
\end{equation}
 where $\left[\cdot ;\cdot \right]$ denotes vector concatenation, $\odot$ denotes dot product and $\mathbf{W}_{o}$ is a learnable parameter. Then we project $\boldsymbol{m}$ to the relation type distribution by MLP and have :
 \begin{equation}
    \begin{aligned}
        p_{\theta_{g}^{rec}}(r_{{e_h},{e_t}}|D_{t-1},\mathcal{G}) =  \operatorname{softmax}\left( \operatorname{MLP}[\boldsymbol{m}]\right).
    \end{aligned}
\end{equation}

To adapt to downstream recommendation and generation subtasks, we simplify relations among the knowledge graph networks into two classes by checking if the entity pair $({e_h},{e_t})$ is connected or not, denoted as $r_{{e_h},{e_t}} = 1$ or $=0$.
The approximate posterior knowledge graph refactor network follows the same process as the prior network but additionally perceives the target entity $I_t$ for approximating the posterior distribution, thereafter, we replace $\boldsymbol{d}_{t-1}$ by $ \boldsymbol{d}_{t-1}^*$ in Eq.~\ref{Eq:info_fusion}. 

Besides, during the training process, we draw $r_{e_h,e_t}$ via gumbel sampling out of the approximate posterior probability distribution $q_{\phi}(r_{e_h,e_t}|I_t,{D_{t-1}},\mathcal{G})$, but in testing we only execute the prior distribution  $p_{\theta_{g}^{rec}}(r_{e_h,e_t}|{D_{t-1}},\mathcal{G})$ and get its max sampling for the CR task.

\subsubsection{Incomplete graph condition pretraining}
\label{subsubsection:pretrain}
In our setting, we assume the latent variable $\mathcal{G}_{D_{t-1}}$ is conditioned on the original incomplete graph $\mathcal{G}$. To utilize information from $\mathcal{G}$, we adopt Kullback-Leibler divergence as the regularization loss $\mathcal{L}_{reg}$ for pretraining the graph encoder and the knowledge graph refactor networks :
\begin{equation}
\label{Eq:regular}
\begin{aligned}
    L_{reg} =
    &\sum^{e_h,e_t}
    \mathbb{KL}
    \left[  p^{org}\left(r_{e_h,e_t}\right)|| p_{\theta_{g}^{rec}}\left(r_{{e_h},{e_t}} |{D_{t-1}},\mathcal{G} \right)  \right]\\
    +
    &\sum^{e_h,e_t}
    \mathbb{KL}
    \left[ p^{org}\left(r_{e_h,e_t}\right)  || q_{\phi}\left(r_{{e_h},{e_t}} |I_t,{D_{t-1}},\mathcal{G} \right) \right],
\end{aligned}
\end{equation}
where $p^{org}(r_{e_h,e_t}) = 1$ if $(e_h,r,e_t)$ exists in $\mathcal{G}$, otherwise $=0$. It forces the graph refactor network to reconstruct relations in $\mathcal{G}$. 
During the fine-tuning stage, we still treat $\mathcal{L}_{reg}$  as a regularization constraint for the graph refactor network to prevent overfitting.

\subsection{Recommender}
\label{subsec:Recommender System}
In this section, we describe how to utilize the 
dialogue-specific subgraphs for the recommendation subtask. Compared with  $\mathcal{G}$,  $\mathcal{G}_{D_{t-1}}$ has been refined according to the dialogue context and contains enhanced relations. Therefore, we aim to learn enhanced user representation $\boldsymbol{e_u}$ as the weighted sum of tail entity representations. Specifically, in the training process we adopt the probability of relation for the type of ``connected'' from the approximate posterior knowledge graph refactor network as the weights of tail entities:
\begin{equation}
\label{Eq:user representation}
\begin{aligned}
    \boldsymbol{e_u} = \sum^{e_h,e_t}
    q_{\phi}(r_{e_h,e_t}= 1|I_t,{D_{t-1}},\mathcal{G})
    \cdot
    \boldsymbol{e_t}.
\end{aligned}
\end{equation}

Since items contained by the 
dialogue-specific subgraphs are more likely to be recommended, we compute the probability of item $I_i$ in the item set to be recommended to the user $u$ with learned user preference $\boldsymbol{e_u}$ as :
\begin{equation}
\begin{aligned}
{P}_{\theta_r^{rec}}(I_i|\mathcal{G}_{D_{t-1}})
=&\alpha \cdot \operatorname{softmax}\left(\boldsymbol{e}_{u}^{\top} \boldsymbol{e}_{i}\right)\\
&+ 
(1-\alpha) \cdot \frac{1}{Z_{rec}} \sum_{e_h}q_{\phi}(r_{e_h,I_i}= 1|I_t,{D_{t-1}},\mathcal{G}),
\end{aligned}
\end{equation}
where $Z_{rec}$ is a normalization factor. $\alpha$ is a weight hyperparameter. If item $I_i$ is not in the tail entity set, we set $q_{\phi}(r_{e_h,I_i}= 1|I_t,{D_{t-1}},\mathcal{G})$ to $0$.
In the testing process, we use probability from the prior knowledge graph refactor network $p_{\theta_g^{rec}}(r_{{e_h},{e_t}  }= 1|{D_{t-1}},\mathcal{G})$ instead.

We target to maximize $P_{\theta_r^{rec}}(I_t|\mathcal{G}_{D_{t-1}})$  while minimizing the divergence between prior and approximate posterior distributions by Eq.~\ref{eq:ELBO}. And we also maintain the regularization constraint $\mathcal{L}_{reg}$.  So the final objective function for the recommendation subtask is:
\begin{equation}
\begin{aligned}
\mathcal{L}^{rec} =  &-\beta \cdot \log \left({P}_{\theta_r^{rec}}(I_t|\mathcal{G}_{D_{t-1}})\right)\\
    & + \gamma \cdot \sum^{e_h,e_t} \mathbb{KL}
    [ q_{\phi} \left( r_{{e_h},{e_t}}  |I_t,{D_{t-1}},\mathcal{G} \right)
    ||p_{\theta_{g}^{rec}}\left(r_{{e_h},{e_t}} |{D_{t-1}},\mathcal{G} \right) 
     ]\\
    & + \lambda \cdot \mathcal{L}_{reg},
\end{aligned}
\end{equation}
where $\beta$, $\gamma$ and $\lambda$ are hyperparameters.

\subsection{Response generator}
\label{subsec:Dialouge System}

Since we have derived a dialogue-specific subgraphs $\mathcal{G}_{D_{t-1}}$ from $D_{t-1}$ and $\mathcal{G}$, we reformulate the probability $P_{\theta^{dia}}(U_{t}|D_{t-1},\mathcal{G})$ in Eq.~\ref{tgt_rec} as $P_{\theta^{dia}}(U_{t}|D_{t-1},\mathcal{G}_{D_{t-1}})$. Here, we introduce our method to accomplish the response generation subtask. 

In the response decoding process, we freeze the parameters of the context encoder and the knowledge graph refactor network, and only use the filtered tail entities inferred to be connected to head entities by the prior network with argmax sampling, as we believe they help generate more consistent and diverse utterances. 
Following ~\citet{DBLP:conf/kdd/ZhouZBZWY20}, our decoder follows a standard transformer decoder architecture but modified to be knowledge enhanced. Specifically, we stack the representations of head entities and tail entities into matrices $\boldsymbol{\mathcal{N}_h}$ and $\boldsymbol{\mathcal{N}_t}$.
After self-attention, we fuse entity representation matrices by conducting two knowledge-based attention layers:
\begin{equation}
\label{eq:trans-dec}
\begin{aligned}
    \boldsymbol{A}_0^l &=\rm MHA[\textbf{C}^{(l-1)},\textbf{C}^{(l-1)},\textbf{C}^{(l-1)}],\\
    \textbf{A}_1^l &=\rm MHA[\textbf{A}_0^l,\boldsymbol{\mathcal{N}_h},\boldsymbol{\mathcal{N}_h}],\\
    \textbf{A}_2^l &=\rm MHA[\textbf{A}_1^l,\boldsymbol{\mathcal{N}_t},\boldsymbol{\mathcal{N}_t}],\\
    \textbf{A}_3^l &=\rm MHA[\textbf{A}_2^l,\textbf{X},\textbf{X}],\\
    \textbf{C}^l &=\rm FFN(\textbf{A}_3^l),
\end{aligned}
\end{equation}
where $\rm MHA[queries,keys,values]$  denotes the multi-head attention function. $\rm FFN(\cdot)$ denotes a fully connected feed-forward network. And $\textbf{X}$ 
is the  representation matrix of dialogue context $D_{t-1}$ obtained from a transformer encoder. Finally, $\textbf{C}^l$ is the representation matrix from the decoder at the $l$-th layer. Besides, we use the copy mechanism~\citep{DBLP:conf/acl/GuLLL16}
, which is also helpful for generating informative utterances.

Suppose the length of ${U}_{i}$ is $|{U}_{i}|$ ,we use a cross-entropy loss to learn the parameters in our generator:
\begin{small}
\begin{equation}
\small
\begin{aligned}
     \mathcal{L}_{dia}=
     -\frac{1}{|{U}_{i}|} \sum_{j=1}^{|{U}_{i}|} \log p_{\theta^{dia}}\left(w_{j}|D_{t-1},\mathcal{G}_{D_{t-1}},w_{0:j-1}\right).
\end{aligned}    
\end{equation}
\end{small}
\section{EXPERIMENTS}
\label{sec:Experimental Setup}
In this section, we conduct experiments to verify the effectiveness of our proposed method. We aim to answer the following research questions:
\begin{enumerate*}[label=\textbf{RQ\arabic*},leftmargin=*,nosep]
\item Does \OurMethod{} outperform state-of-the-art CR methods in terms of recommendation and response generation?
\item How does each component of \OurMethod{} contribute to its overall performance? 
\item Can \OurMethod{} generate reasonable and diverse responses based on incomplete KGs?
\end{enumerate*}

\subsection{Datasets}

In the experiments, we use two widely-applied \ac{CRSs} datasets to verify the effectiveness of \OurMethod{}, including REDIAL~\citep{DBLP:conf/nips/LiKSMCP18} and TG-REDIAL ~\citep{DBLP:conf/coling/ZhouZZWW20}. They are both composed of two-party dialogues between a user and a recommender in the movie domain.

 \noindent
 \textbf{REDIAL dataset.} The REDIAL dataset~\citep{DBLP:conf/nips/LiKSMCP18} is a collection of English conversational recommendation sessions built manually by crowd-sourcing workers from Amazon Mechanical Turk platform. It is widely used in CRSs~\citep{DBLP:conf/wsdm/ZhouZZWJ022,DBLP:conf/emnlp/MaTH21,DBLP:conf/emnlp/ZhouWHH21,DBLP:conf/kdd/ZhouZBZWY20,DBLP:conf/emnlp/ChenLZDCYT19}. It contains 10,006 sessions, consisting of 182,150 utterances related to 51,699 movies. 
 
\noindent
\textbf{TG-REDIAL dataset.} The TG-REDIAL is a Chinese conversational recommendation dataset constructed in a semi-automatic topic-guided way to enforce natural semantic transitions towards recommendation. It is composed of 10,000 sessions of 129,392 utterances related to 33,834 movies.
 
\subsection{Baselines}
In the experiments, we evaluate \OurMethod{} on 
recommendation and response generation. We compare our approach with not only existing CRSs but also representative recommendation and response generation approaches:
\begin{enumerate*}[label={(\arabic*)}]
 \item \textbf{Recommendation approaches}: To evaluate the performance of \OurMethod{} in recommendation subtask, we use 
 Popularity, 
 TextCNN~\citep{DBLP:conf/emnlp/Kim14},
 and BERT~\citep{DBLP:conf/naacl/DevlinCLT19} as baselines. 
 Popularity ranks the items in the corpus based on past recommendation frequencies. 
 TextCNN employs a CNN-based approach to extract textual information from contextual utterances. 
 BERT is a pretrained language model for recommendation that directly encodes the concatenated historical utterances.
 \item \textbf{Conversational approach}: To assess the performance of response generation, we consider Transformer~\citep{DBLP:conf/nips/VaswaniSPUJGKP17}. It uses a Transformer-based encoder and decoder framework to provide appropriate replies.
 \item \textbf{Knowledge grounded conversational approach}: To evaluate the performance of \OurMethod{} in recommendation and response generation, we use PostKS~\citep{DBLP:conf/ijcai/LianXWPW19} as a baseline. It uses dialogue context and responses to infer posterior item distributions and generates responses.
 \item \textbf{Conversational recommender systems}: To evaluate the performance of \OurMethod{} in \ac{CR} task, we use 
 REDIAL~\citep{DBLP:conf/nips/LiKSMCP18}, 
 KBRD~\citep{DBLP:conf/emnlp/ChenLZDCYT19}, 
 MGCG~\citep{DBLP:conf/acl/LiuWNWCL20},
 KGSF~\citep{DBLP:conf/kdd/ZhouZBZWY20},
 C{\({^2}\)}-CRS~\citep{DBLP:conf/wsdm/ZhouZZWJ022} and 
 UniCRS~\citep{DBLP:conf/kdd/WangZWZ22} as baselines. 
 REDIAL is a benchmark model of the REDIAL dataset. It comprises of a HRED-based\citep{DBLP:conf/aaai/SerbanSLCPCB17} dialog generation module, an Autoencoder-based~\citep{DBLP:conf/www/SedhainMSX15} recommender module, and a sentiment analysis module.
 KBRD uses \ac{KGs} to improve the semantics fusion for the recommender and dialouge systems. It uses  Transformer~\citep{DBLP:conf/nips/VaswaniSPUJGKP17} to provide replies with enhanced modeling of word weights.
MGCG predicts the target items and generates responses using a CNN-based multi-task classifier and a GRUs-based generator.
 KGSF incorporates both word-oriented and entity-oriented \ac{KGs} to enhance the user representations and generate responses with Transformer frameworks.
 C{\({^2}\)}-CRS proposes a coarse-to-fine contrastive learning framework to align the associated multi-type semantic units in CRSs. For a fair comparison, we removed the review information from the C{\({^2}\)}-CRS. 
 UniCRS unifies the recommendation and response generation subtasks into the prompt learning paradigm based on a pretrained language model. Since UniCRS uses a language-specific pretrained model, we only compare with it on the REDIAL dataset.
\end{enumerate*}

\subsection{Evaluation metrics}
To evaluate the effectiveness of our methods, we utilize both automatic evaluation and human evaluation.

\noindent
\textbf{Automatic evaluation}. To evaluate the performance on the recommendation subtask,  following~\citep{DBLP:conf/emnlp/ChenLZDCYT19}, we adopt Recall@$m$($m = 1,10,50$) which measures whether the ground-truth items are in the top-$m$ positions of the recommendation list. To assess the quality of dialogue generation, we adopt Distinct-$n$($n = 3,4$) for diversity. Meanwhile, we apply ROUGE-$n$($n = 1,2,l$)  for recall of generated n-grams on word-level comparison with the ground-truth responses.
 
\noindent
\textbf{Human evaluation}. For CRSs, it's crucial that the system can provide informative and fluent replies for personalized recommendation. Hence, we recruit three annotators to assess the results of all dialogue models in two aspects, i.e., fluency and informativeness. Fluency evaluates if the generated response is smooth; informativeness assesses whether the system contributes interesting information. The score goes from 0 to 2. The final performance is computed using the average ratings of the three annotators. We abbreviate Fluency and Informativeness as Flu and Inf in Table ~\ref{tab:generation result}.

\subsection{Implementation details}
We implement \OurMethod{} with PyTorch. 
The dimension of entity representations is 128, and the embedding size of words is 768. The number of tail entities $k$ is 40. 
Hyperparameter $\alpha$ is 0.1, $\beta$, $\gamma$ and $\lambda$ are set to 1, 10 and 0.0025, respectively. The number of RGCN layers is set to 1. We use the Adam optimizer ~\citep{DBLP:journals/corr/KingmaB14}. In the recommender system, we set the learning rate for the BERT encoder and RGCN as $1e-5$ and $5e-4$, respectively. The learning rate for the other components are set as $1e-3$. For the response generator we adopt the warm-up learning rate schedule. The learning rate is initialized to 0.5 and warm-up step is set to 2000. The word vocabulary size for the dialogue system is 30000 on TG-REDIAL and 23929 on REDIAL. 
Following~\citet{DBLP:conf/wsdm/ZhouZZWJ022}, we begin with the first sentence one by one to come up with recommendation or reply to utterances for each conversation. 

\section{Experimental ResultS}
\label{sec:Experimental Result and Analysis}

\begin{table}
\small
\caption{Automatic evaluation of recommendation on TG-REDIAL and REDIAL datasets. Boldface indicates the best result. Significant improvements over best baseline results are marked with * (t-test, $p \textless 0.05)$.}
\label{tab:recommendation results}
\begin{tabular}{l@{~}cccc ccc}
\toprule
\multirow{3}{*}{Model} & \multicolumn{3}{c}{\textbf{REDIAL}} & \multicolumn{3}{c}{\textbf{TG-REDIAL}}\\
\cmidrule(r){2-4} \cmidrule(l){5-7}  
& \multicolumn{3}{c}{Recall} & \multicolumn{3}{c}{Recall} \\
\cmidrule(r){2-4} \cmidrule(r){5-7}
& @1     & @10   & @50  
& @1     &  @10  & @50 \\ 
\midrule
Popularity &  0.011 &  0.055 &  0.181 &  0.000 &  0.003 &  0.015\\ 
TextCNN & 0.013 &  0.067 &  0.188 &  0.003 &  0.010 &  0.023 \\
BERT &  0.014 &  0.117 &  0.191 &  0.003 &  0.016 &  0.029 \\
PostKS & 0.019 & 0.122 & 0.236  & 0.003 & 0.017  & 0.045\\
ReDial &  0.023 &  0.141 &  0.320 &  0.000 &  0.003 &  0.015 \\
MGCG &  0.027 &  0.121 &  0.264 &  0.004 &  0.028 &  0.068 \\
KBRD &  0.031 &  0.159 &  0.335 &  0.005 &  0.032 &  0.077 \\
KGSF &  0.036 &  0.177 &  0.363 &  0.005 &  0.030 &  0.073 \\
C{\({^2}\)}-CRS  &  0.050 &  0.218 &  0.395 &  0.005 &  0.031 & 0.077 \\
UniCRS &  {0.051} &  {0.224} &  \textbf{0.428} &{---}   &{---}   &{---}  \\
\midrule 
\OurMethod{}  & \textbf{0.057}\rlap{$^\ast$}  & \textbf{0.251}\rlap{$^\ast$}  & 0.416  & \textbf{0.005}  &
\textbf{0.032}   & \textbf{0.081}\rlap{$^\ast$}\\ 
\bottomrule
\end{tabular}
\end{table}

\subsection{Performance comparison (RQ1)}
\subsubsection{\textbf{Comparison on recommendation}}

\
\newline
\noindent Table~\ref{tab:recommendation results} shows the result of recommendation performance. We observe that \OurMethod{} outperforms all the baselines on both datasets in terms of most metrics. For the REDIAL dataset, \OurMethod{} achieves a significant 11.8\% and 12.1\% improvement over UniCRS in terms of Recall@1 and Recall@10, respectively. 
For TG-REIDAL datasest, \OurMethod{} achieves an increase of 3.2\% and 5.2\% over C{\({^2}\)}-CRS on Recall@10 and Recall@50, respectively.
It proves \OurMethod{} provides better recommendation after aggregating the evidence from \ac{KGs} and the dialogue corpus. Moreover, \OurMethod{} refines knowledge of \ac{KGs} according to specific dialogue context, which allows the model to obtain higher accuracy. 
However, we find that \OurMethod{} is slightly deficient in comparison with UniCRS for the REDIAL dataset on Recall@50. We consider it is because UniCRS additionally exploits the implicit knowledge within its pretrained language model.

For baselines, we find that typical context-aware recommendation models (such as TextCNN and BERT) outperform Popularity, indicating that contextual information can be more beneficial for recommendation than general item popularity.
PostKS outperforms context-aware recommendation models since it takes movies as knowledge and utilizes variational Bayesian inference for more precise knowledge selection.
For \ac{CRSs} methods, they generally perform  better than traditional context-aware and knowledge-grounded conversational approach since they bridge dialogue and recommendation to promote both subtasks. 
\ac{KG}-aware CR approaches (such as KBRD, KGSF, C{\({^2}\)}-CRS, and UniCRS) significantly outperform other baselines, implying that the \ac{KGs} are critical for the recommendation.
\begin{table*}[t]
\small
\centering
\setlength\tabcolsep{6.0pt}
\caption{Automatic and human evaluation of response generation on REDIAL and TG-REDIAL datasets. 
Boldface indicates the best result. Significant improvements over best baseline results are marked with * (t-test, $p \textless 0.05)$.}

\label{tab:generation result}
\begin{tabular}{l@{~}ccccccc ccccccc}
\toprule
\multirow{3}{*}{Model} & \multicolumn{7}{c}{\textbf{REDIAL}} & \multicolumn{7}{c}{\textbf{TG-REDIAL}}\\
\cmidrule(r){2-8} \cmidrule(l){9-15}  
& \multicolumn{2}{c}{Distinct} & \multicolumn{3}{c}{ROUGE} & \multicolumn{2}{c}{Human} & \multicolumn{2}{c}{Distinct} &\multicolumn{3}{c}{ROUGE} & \multicolumn{2}{c}{Human}\\
\cmidrule(r){2-3} \cmidrule(r){4-6} \cmidrule(r){7-8} \cmidrule(r){9-10} \cmidrule{11-13} \cmidrule{14-15}
& @3     & @4  & @1   & @2   & @$l$   & Flu  & Info   
&  @3  & @4 & @1   & @2   & @$l$  & Flu  & Info\\ 
\midrule
Transformer & 0.128 & 0.196 &0.135  &0.024  &0.130  & 0.790 & 0.604 & 0.106 & 0.187 & 0.307 & 0.064 & 0.261   &0.802  &0.781  \\
PostKS & 0.126 & 0.224 & 0.124   & 0.014   & 0.113  & 0.647  & 0.698  & 0.147  & 0.250 & 0.264  & 0.046  &0.233  & 0.767  & 0.837  \\
ReDial &  0.135 & 0.188 & 0.139   &0.012   &0.130  & 0.763  & 0.710  &  0.153 & 0.226 &  0.303 &  0.032 & 0.269  & 0.629 & 0.733 \\
MGCG & 0.167   & 0.252  & 0.101  & 0.016  & 0.092 & 0.614  & 0.926  &  0.173 &  0.253 &  0.247 &  0.041 &  0.221  &0.781  &1.330  \\
KBRD &  0.168 &  0.245 & 0.112  & 0.021  & 0.108  & 0.697  & 0.986  &  0.166 &  0.268 &  0.328 &  0.056 &  0.279  &0.845  &1.244  \\
KGSF &  0.227 &  0.308 & 0.125  & 0.025   &0.120
   & 0.957  & 1.331  &  0.208 &  0.317 &  0.293 &  0.048 &  0.240  &1.011  &1.365  \\
C{\({^2}\)}-CRS &  0.227 &  0.339 &  0.121 &  0.020 &  0.115 & 1.200   &  1.556   &  0.268 &  0.365 &  0.270 &  0.046 &  0.228  &1.314  &1.473  \\ 
UniCRS & 0.231  & 0.359 & 0.103 & \textbf{0.032} & 0.101  & 1.297   & 1.209    & {---}  & {---}  & {---}  & {---} & {---}   & {---} & {---} \\ 
\midrule
\OurMethod{} & \textbf{0.292}\rlap{*} & \textbf{0.482}\rlap{*} &  \textbf{0.144}\rlap{*} & 0.026 & \textbf{0.137}\rlap{*}  &  \textbf{1.302} & \textbf{1.724}\rlap{*} & \textbf{0.408}\rlap{*} & \textbf{0.596}\rlap{*}
& \textbf{0.329}\rlap{*}  & \textbf{0.091}\rlap{*}  & \textbf{0.289}\rlap{*} &  \textbf{1.515}\rlap{*} & \textbf{1.650}\rlap{*}\\
\bottomrule
\end{tabular}

\end{table*}

\subsubsection{\textbf{Comparison on response generation}}

\ 
\newline
\noindent \textbf{Automatic evaluation}.
Table~\ref{tab:generation result} shows the performance comparison of response generation. We see that \OurMethod{} performs better than all baselines in terms of most automatic metrics on both datasets. The reason is \OurMethod{} refines knowledge conditioned on specific context to reduce extraneous information, which results in generating responses closer to the ground truth. Besides, \OurMethod{} also greatly improves the diversity by inferring missing relations.

Among all the baselines, we find that UniCRS and C{\({^2}\)}-CRS perform the best in terms of diversity on the REDIAL and TG-REDIAL datasets. One possible reason is that UniCRS uses pretrained generative models and C{\({^2}\)}-CRS applies an instance weighting mechanism, which improve the generation performance. 
However, both UniCRS and C{\({^2}\)}-CRS sacrifice the match with truth responses and get lower ROUGE values. 
In summary, we conclude that inferring dialogue-specific subgraphs is helpful for generating consistent and diverse responses.

\noindent
\textbf{Human evaluation}.
Table ~\ref{tab:generation result} also lists the human evaluation result of response generation. We find that \OurMethod{} achieves an increase of 8.5 (15.3)\% and 10.8 (12.0)\% over C{\({^2}\)}-CRS on REDIAL (TG-REDIAL) dataset in terms of fluency and informativeness, which indicates that \OurMethod{} is able to generate more informative utterances by incorporating inferred entity representations and the copy mechanism. Besides,  dynamic reasoning on \ac{KGs} reduces the semantic gap between context and knowledge, resulting in more fluent responses. 
In addition, we find C{\({^2}\)}-CRS is outstanding among all baselines by using contrastive learning to facilitate semantic fusion of \ac{KGs} and dialogues, except for the Fluency on the REDIAL dataset. In the process of manual annotation, we observe that compared with C{\({^2}\)}-CRS, UniCRS tends to generate fluency but templated responses, which greatly reduces the user's experience.

To conclude, VRICR outperforms state-of-the-art \ac{CR} methods in terms of both recommendation and response generation.

\vspace*{-2mm} 
\subsection{Ablation study (RQ2)}

\begin{table}[t]
\centering
\setlength\tabcolsep{10pt}
\caption{
Results of ablation studies for the recommendation subtask in REDIAL. Boldface indicates the best result. Significant improvements are marked with $^\ast$ (t-test, $p < 0.05$).}
\label{tab:ablation study}
\begin{tabular}{lccc}
\toprule
 Model & Recall@1 & Recall@10 & Recall@50\\
\midrule
\OurMethod{}$\backslash${P}  & 0.016 & 0.083 & 0.217 \\ 
\OurMethod{}$\backslash${C}  & 0.055 & 0.250 & 0.406 \\
\OurMethod{}$\backslash${I}  & 0.041 & 0.164 & 0.228  \\
\midrule 
\OurMethod{}  & \textbf{0.057}\rlap{$^\ast$}  & \textbf{0.251}\rlap{$^\ast$}  & \textbf{0.416}\rlap{$^\ast$}  \\ 
\bottomrule
\end{tabular}
\end{table}

We conduct an ablation study on the REDIAL dataset to examine where the improvements of \OurMethod{} come from. The results are shown in Table ~\ref{tab:ablation study}. 
We considered three variants of the model, including: 
(1)\OurMethod{}$\backslash${P} removes the pretraining processing in (sec~\ref{subsubsection:pretrain}); (2)\OurMethod{}$\backslash${C} removes dialogue contextual information in knowledge graph refactor network; and (3)\OurMethod{}$\backslash${I} freezes parameters of refactor network during fine-tuning, thereby removes dynamic relations reasoning.

We see that all components are helpful for the recommendation because the performance drops in all three variants. Specifically, we find that \OurMethod{}$\backslash${P} leads to the largest performance decrease as the randomness in the relation reasoning process hinders the simultaneous completion of reasoning and recommendation. The pretraining process effectively learns entity representations that are more adaptable to the recommendation {subtask}. \OurMethod{}$\backslash${C} leads to a performance decrease. It shows that perceiving dialogue context, \OurMethod{} constructs dialogue-specific subgraphs for more accurate recommendation. Finally, \OurMethod{}$\backslash${I} combining original \ac{KGs} without dynamic relations reasoning yields a worse performance than \OurMethod{}. The major reason is that the original \ac{KG} contains noise and extraneous information.

\vspace*{-2mm}
\subsection{Case study (RQ3)}

\begin{table}[t]
  \caption{One case extracted from the REDIAL dataset.}
  \label{tab:Case Study}
  \begin{tabularx}{\linewidth}{lX}
  \toprule
  
  \multirow{3}{*}{\rotatebox[origin=c]{90}{\textbf{Context}}} 
   & 
   {\textbf{s1(recommender)}: What type of movie do you enjoy watching ?}\\
   ~ & {\textbf{s2(user)}: I was thinking of putting on a classic movie and something along the lines of \textit{\textbf{North by Northwest}}. It was a thrilling movie.}\\
   \midrule
  \multirow{9}{*}{\rotatebox[origin=lt]{90}{\textbf{Response}}} 
   &
    {\textbf{C{\({^2}\)}-CRS}: Have you seen the movie \textit{\textbf{Gone with the Wind}}?}\\
   \cmidrule(r){2-2} 
   ~ & {\textbf{UniCRS}: That was a great movie!} \\ 
   \cmidrule(r){2-2} 
   & {\textbf{\OurMethod{}}: I know of that one , I liked it a lot. How about \emph{\textbf{The Day of the Jackal}} ?}\\
   ~ &{\emph{inferred linked entities }: \textit{The Day of the Jackal; Marnie; Shallow Grave}.} \\ 
   
    \bottomrule
  \end{tabularx}
\end{table}
To examine if \OurMethod{} generates reasonable and diverse responses based on incomplete \ac{KGs}, we randomly sample a conversation generated by three models (i.e., VRICR, C$^2$-CRS, and UniCRS ) on the REDIAL dataset. 
As shown in Table ~\ref{tab:Case Study}, in the third utterance, awarding of essential context {``classic'' and ``thrilling''}, \OurMethod{} establishes the relations between the historically interacted entity ``North by Northwes'' and {``The Day of the Jacka'', ``  Marnie'', ``Shallow Grave''}, which does not exist in the original \ac{KGs}. All the inferred linked entities are classic movies produced in the twentieth century, and we find they all belong to the thriller category. It proves that \OurMethod{} is capable of inferring  reasonable {and diverse potential} missing relations. Then \OurMethod{} uses one of inferred relations to generate a response. While C{\({^2}\)}-CRS ignores the thriller attributes and UniCRS is more likely to generate generic responses, leading to a decline in user experience. \OurMethod{} can incorporate dialogue corpus to enhance the incomplete \ac{KGs} and refine knowledge conditioned on specific dialogue context, which increases reasonableness and diversity of responses.

\vspace*{-2mm}
\section{CONCLUSION AND FUTURE WORK}
\label{sec:CONCLUSION}
The main contribution of this paper is to explicitly tackle  the problem of incompleteness, sparsity, and noise in \ac{KGs} for better conversational recommendation. We have incorporated the evidence from both dialogue corpus and KGs to infer the dialogue-specific subgraphs of \ac{KG} and perform adaptive knowledge selection for recommendation. To this end, we have proposed \OurMethod{}, which regards the 
dialogue-specific subgraphs as latent variables with categorical priors for adaptive knowledge graphs refactor. As a result, \OurMethod{} not only infers the missing relations of incomplete \ac{KGs} but also dynamically selects relevant knowledge. We have conducted experiments on two benchmark datasets. Experimental results verify the effectiveness of our proposed method. 

The limitation of our work is that \OurMethod{} does not fuse recommendation and response generation within an end-to-end fashion. Therefore, it hinders the mutual promotion of recommendation and response generation. 
For future work, one promising direction is to introduce dynamic variational reasoning in response generation to improve the interpretability of utterances.

\section*{Reproducibility}
This work uses publicly available data. To facilitate reproducibility of the results reported in this paper, the code used is available at \url{https://github.com/zxd-octopus/VRICR}.

\begin{acks}
This work was supported by the Natural Science Foundation of China (62072279,62272274,61902219, 61972234, 62102234, 62202271), the Natural Science Foundation of Shandong Province (ZR2021QF129), the Key Scientific and Technological Innovation Program of Shandong Province (2019JZZY010129), Shandong University multidisciplinary research and innovation team of young scholars (No. 2020QNQT017), Meituan. 
\end{acks}



\clearpage
\bibliographystyle{ACM-Reference-Format}
\balance{}
\bibliography{main}
\clearpage
\appendix

\section{APPENDIX}
\label{formular}
\subsection{Evidence Lower Bound Derivation}

The objective function of the recommendation subtask is:
\begin{equation}
\begin{aligned}
& P\left(I_{t}\mid D_{t-1},\mathcal{G}\right) \\
&= \sum_{\mathcal{G}_{D_{t-1}}} P\left(I_{t}\mid D_{t-1},\mathcal{G},\mathcal{G}_{D_{t-1}}\right) \\
&=\sum_{\mathcal{G}_{D_{t-1}}}P(I_{t}|\mathcal{G}_{D_{t-1}})\cdot P(\mathcal{G}_{D_{t-1}}|D_{t-1},\mathcal{G}),
\end{aligned}
\end{equation}

To derive the ELBO, we deform the objective function:

\begin{equation}
\begin{aligned}
&\log P\left(I_{t}\mid D_{t-1},\mathcal{G}\right)\\
&=\log \frac{P\left(I_{t}, \mathcal{G}_{D_{t-1}} \mid D_{t-1},\mathcal{G}\right)}
{P\left(\mathcal{G}_{D_{t-1}} \mid I_{t},D_{t-1}, \mathcal{G}\right)}\\
&=\log P\left(I_{t}, \mathcal{G}_{D_{t-1}} \mid D_{t-1},\mathcal{G}\right)
-\log P\left(\mathcal{G}_{D_{t-1}} \mid I_{t},D_{t-1}, \mathcal{G}\right),
\end{aligned}
\end{equation}

We introduce a new distribution $Q\left(\mathcal{G}_{D_{t-1}} \mid I_{t}, D_{t-1}, \mathcal{G}\right)$:

\begin{equation}
\begin{aligned}
\label{eq:elbo_1}
&\log P\left(I_{t}\mid D_{t-1},\mathcal{G}\right)\\
&=\log \frac{P\left(I_{t}, \mathcal{G}_{D_{t-1}} \mid D_{t-1}, \mathcal{G}\right)}
{Q\left(\mathcal{G}_{D_{t-1}} \mid I_{t}, D_{t-1}, \mathcal{G}\right)}
-\log \frac{P\left(\mathcal{G}_{D_{t-1}} \mid I_{t},D_{t-1}, \mathcal{G}\right)}
{Q\left(\mathcal{G}_{D_{t-1}} \mid  I_{t},D_{t-1}, \mathcal{G}\right)},
\end{aligned}
\end{equation}

The left and right sides of the Eq.~\ref{eq:elbo_1} are simultaneously integrated over $\mathcal{G}_{D_{t-1}}$  and multiplied by $Q\left(\mathcal{G}_{D_{t-1}} \mid I_{t},D_{t-1}, \mathcal{G}\right)$:

\begin{equation}
\begin{aligned}
&\sum_{\mathcal{G}_{D_{t-1}}}
Q\left(\mathcal{G}_{D_{t-1}} \mid I_{t},D_{t-1}, \mathcal{G}\right)
\log P\left(I_{t}\mid D_{t-1},\mathcal{G}\right)\\
&=\sum_{\mathcal{G}_{D_{t-1}}} Q\left(\mathcal{G}_{D_{t-1}} \mid I_{t},D_{t-1}, \mathcal{G}\right)\times \log \frac{P\left(I_{t}, \mathcal{G}_{D_{t-1}} \mid D_{t-1}, \mathcal{G}\right)}{Q\left(\mathcal{G}_{D_{t-1}} \mid I_{t},D_{t-1}, \mathcal{G}\right)}\\
&-\sum_{\mathcal{G}_{D_{t-1}}} Q\left(\mathcal{G}_{D_{t-1}} \mid I_{t},D_{t-1}, \mathcal{G}\right) \times \log \frac{P\left(\mathcal{G}_{D_{t-1}} \mid I_{t}, D_{t-1}, \mathcal{G}\right)}{Q\left(\mathcal{G}_{D_{t-1}} \mid I_{t},D_{t-1}, \mathcal{G}\right)},\\\\
\end{aligned}
\end{equation}

Because:
\begin{equation}
\begin{aligned}
&\sum_{\mathcal{G}_{D_{t-1}}} Q\left(\mathcal{G}_{D_{t-1}} \mid I_{t},D_{t-1}, \mathcal{G}\right) \times \log \frac{P\left(\mathcal{G}_{D_{t-1}} \mid I_{t}, D_{t-1}, \mathcal{G}\right)}{Q\left(\mathcal{G}_{D_{t-1}} \mid I_{t},D_{t-1}, \mathcal{G}\right)}\\
&=-\mathbb{KL}\left[Q\left(\mathcal{G}_{D_{t-1}} \mid I_{t},D_{t-1}, \mathcal{G}\right)||P\left(\mathcal{G}_{D_{t-1}} \mid I_{t}, D_{t-1}, \mathcal{G}\right)\right]\\
&\leq 0,\\\\
\end{aligned}
\end{equation}

So we have:
\begin{equation}
\begin{aligned}
\label{eq:elbo_2}
&\log P\left(I_{t} \mid D_{t-1}, \mathcal{G}\right) \\
&\geqslant E L B O\\
 &=\sum_{\mathcal{G}_{D_{t-1}}} Q\left(\mathcal{G}_{D_{t-1}} \mid I_{t},D_{t-1}, \mathcal{G}\right) \times \log \frac{P\left(I_{t}, \mathcal{G}_{D_{t-1}} \mid D_{t-1}, \mathcal{G}\right)}{Q\left(\mathcal{G}_{D_{t-1}} \mid I_{t},D_{t-1}, \mathcal{G}\right)}\\
&=\sum_{\mathcal{G}_{D_{t-1}}} 
Q\left(\mathcal{G}_{D_{t-1}} \mid I_{t},D_{t-1}, \mathcal{G}\right)
\times \log \frac
{P\left(I_{t} \mid \mathcal{G}_{D_{t-1}}\right) \times P\left(\mathcal{G}_{D_{t-1}} \mid D_{t-1},  \mathcal{G}^\text {inc }\right)}
{Q\left(\mathcal{G}_{D_{t-1}} \mid  I_{t}, D_{t-1}, \mathcal{G}\right)}\\
&=\sum_{\mathcal{G}_{D_{t-1}}} Q\left(\mathcal{G}_{D_{t-1}} \mid I_{t},D_{t-1}, \mathcal{G}\right) \times 
\log P\left(I_{t} \mid \mathcal{G}_{D_{t-1}} \right)\\
&+\sum_{\mathcal{G}_{D_{t-1}}} Q\left(\mathcal{G}_{D_{t-1}} \mid I_{t},D_{t-1}, \mathcal{G}\right) \times \log \frac{P\left( { \mathcal{G}^\text{enh }} \mid D_{t-1},  { \mathcal{G} }\right)}{Q\left(\mathcal{G}_{D_{t-1}} \mid I_{t},D_{t-1}, \mathcal{G}\right)}\\
&=\textcircled{1} + \textcircled{2},\\
\end{aligned}
\end{equation}

At the $t$-th turn, $\mathcal{G}_{D_{t-1}}$ is derived depending on dialogue context $D_{t-1}$ and incomplete graph $\mathcal{G}$. 
For simplicity, we further assume that relations of entity pair $(e_h, e_t)$ are  independent variables, so we reformulate distribution of $P(\mathcal{G}_{D_{t-1}}|D_{t-1},\mathcal{G})$ and $Q(\mathcal{G}_{D_{t-1}}|I_t, D_{t-1},\mathcal{G})$ as:

\begin{equation}
\begin{aligned}
P(\mathcal{G}_{D_{t-1}}|D_{t-1},\mathcal{G}) &= \prod^{e_h ,e_t }p(r_{{e_h},{e_t}}|D_{t-1},\mathcal{G});\\
Q(\mathcal{G}_{D_{t-1}}|I_t, D_{t-1},\mathcal{G}) &= \prod^{e_h ,e_t}q(r_{{e_h},{e_t}}|I_t, D_{t-1},\mathcal{G})\\
\end{aligned}
\end{equation}

We reformulate Eq.~\ref{eq:elbo_2} according to the assumptions mentioned above:
\begin{equation}
\begin{aligned}
&\textcircled{1}\\
&=\sum_{\mathcal{G}_{D_{t-1}}} Q\left(\mathcal{G}_{D_{t-1}} \mid I_{t},D_{t-1}, \mathcal{G}\right) \times \log P\left(I_{t} \mid \mathcal{G}_{D_{t-1}}\right)\\
&=\mathbb{E}_{\prod^{e_h ,e_t}q\left(r_{{e_h},{e_t}} \mid I_{t}, D_{t-1},  { \mathcal{G} }\right)}\left[\log P\left(I_{t} \mid \mathcal{G}_{D_{t-1}}\right)\right]\\
\end{aligned}
\end{equation}
\begin{equation}
\begin{aligned}
& \textcircled{2}\\
&=\sum_{r_{{e_h},{e_t}} \in R}^{e_h ,e_t } \prod^{e_h ,e_t } q\left(r_{{e_h},{e_t}} \mid I_{t},D_{t-1}, \mathcal{G} \right) \times \log \frac
{\prod^{e_h ,e_t }p\left(r_{{e_h},{e_t}} \mid D_{t-1}, \mathcal{G}\right)}
{\prod^{e_h ,e_t }q\left(r_{{e_h},{e_t}} \mid I_{t},D_{t-1}, \mathcal{G}\right)} \\
&=\sum_{e_i ,e_j} \sum_{r_{{e_h},{e_t}} \in R}^{e_h ,e_t} \prod^{e_h ,e_t } q\left(r_{{e_h},{e_t}} \mid I_{t},D_{t-1}, \mathcal{G} \right) \times \log \frac
{p\left(r_{{e_i},{e_j}} \mid D_{t-1}, \mathcal{G}\right)}
{q\left(r_{{e_i},{e_j}} \mid I_{t},D_{t-1}, \mathcal{G}\right)} \\
&=\sum_{e_i ,e_j} 1 \times \sum_{r_{{e_i},{e_j}} \in R} q\left(r_{{e_i},{e_j}} \mid I_{t},D_{t-1}, \mathcal{G} \right) \times \log \frac
{p\left(r_{{e_i},{e_j}} \mid D_{t-1}, \mathcal{G}\right)}
{q\left(r_{{e_i},{e_j}} \mid I_{t},D_{t-1}, \mathcal{G}\right)} \\
&=-\sum_{e_i ,e_j} \mathbb{K L}\left[ q\left(r_{{e_i},{e_j}} \mid I_{t},D_{t-1}, \mathcal{G}^{\text {inc}} \right)||p\left(r_{{e_i},{e_j}} \mid D_{t-1}, \mathcal{G}\right)\right]\\
&=-\sum^{e_h ,e_t} \mathbb{K L}\left[q\left(r_{{e_h},{e_t}} \mid I_{t},D_{t-1}, \mathcal{G}^{\text {inc}} \right)|| p\left(r_{{e_h},{e_t}} \mid D_{t-1}, \mathcal{G}\right) \right]\\
\end{aligned}
\end{equation}

Finally we get the ELBO:
\begin{equation}
\label{eq:ELBO_}
\begin{aligned}
     && &ELBO\\
    &&\ &= \mathbb{E}_{\prod^{e_h ,e_t} q(r_{e_h,e_t}|I_{t}, D_{t-1},  { \mathcal{G} } )}
    \left[
    P
        \left( {I}_{t}|\mathcal{G}_{D_{t-1}} \right) 
    \right]\\
    &&\  &\ \  \ \ - \sum^{e_h,e_t} \mathbb{KL}
    \left[q \left( r_{{e_h},{e_t}} |I_{t}, D_{t-1},  { \mathcal{G} }  \right) || p\left(r_{{e_h},{e_t}}|D_{t-1},  { \mathcal{G} }  \right)  \right], \\
\end{aligned}
\end{equation}

\end{document}